\documentclass{sig-alternate}

\usepackage{graphicx}
\usepackage{subfigure}
\usepackage{cite}

\newtheorem{ex}{Example}

\newcommand{\vertical}{temporal}
\newcommand{\Vertical}{Temporal}
\newcommand{\VerticalTitle}{TEMPORAL}
\newcommand{\horizontal}{genotype}
\newcommand{\Horizontal}{Genotype}
\newcommand{\HorizontalTitle}{GENOTYPE}

\begin{document}
\conferenceinfo{GECCO'07,} {July 7--11, 2007, London, England, United Kingdom.} 
\CopyrightYear{2007} 
\crdata{978-1-59593-697-4/07/0007}

\title{A Doubly Distributed Genetic Algorithm\\for Network Coding}

\numberofauthors{1}
\author{
\alignauthor
Minkyu Kim\raisebox{6pt}{$\ast$}, Varun Aggarwal\raisebox{6pt}{$\dagger$}, Una-May O'Reilly\raisebox{6pt}{$\dagger$}, Muriel M$\acute{\mbox{e}}$dard\raisebox{6pt}{$\ast$}\\
    \affaddr{\raisebox{4pt}{$\ast$}Laboratory for Information and Decision Systems}\\
    \affaddr{\raisebox{2pt}{$\dagger$}Computer Science and Artificial Intelligence Laboratory}\\
    \affaddr{Massachusetts Institute of Technology}\\
    \affaddr{Cambridge, MA 02139, USA}\\
        \email{\{minkyu@, varun\_ag@, unamay@csail., medard@\}mit.edu}
}
\maketitle

\begin{abstract}
We present a genetic algorithm which is distributed in two novel ways: along \horizontal{} and \vertical{} axes. Our algorithm first distributes, for every member of the population, a subset of the genotype to each network node, rather than a subset of the population to each. This \horizontal{} distribution is shown to offer a significant gain in running time. Then, for efficient use of the computational resources in the network, our algorithm divides the candidate solutions into pipelined sets and thus the distribution is in the temporal domain, rather that in the spatial domain. This \vertical{} distribution may lead to temporal inconsistency in selection and replacement, however our experiments yield better efficiency in terms of the time to convergence without incurring significant penalties.
\end{abstract}

\vspace{1mm}
\noindent
{\bf Categories and Subject Descriptors:} C.2.1 {[Computer-Communication Networks]}: {Network Architecture and Design}

\vspace{1mm}
\noindent
{\bf General Terms:} Algorithms

\vspace{1mm}


\section{Introduction}
\label{sec:Intro}

We present a GA which is distributed in two novel ways: along \horizontal{} and \vertical{} axes. In contrast to a conventional GA spatially distributed on the population axis, our doubly distributed algorithm first distributes, \emph{for every member of the population}, a \emph{subset of the genotype} to each network node rather than a subset of the population to each. The motivation for this \horizontal{} axis of distribution is to distribute the fitness evaluation steps of the Network Coding GA (NCGA) \cite{KAME06} which relies on network codes generated randomly and in a decentralized manner. Self-referentially, the GA solving the network coding problem must be embedded in the same network for which it is searching for the optimal coding. With just this axis of distribution, the distributed NCGA equals the performance of the (centralized) NCGA in terms of solution quality. However, as experiments herein suggest, it can lead to a significant gain in running time.

The motivation for the second axis of distribution is to maximize the efficient use of the computational nodes in the network by minimizing their idle duration during the GA search. Along this second, \vertical{} axis of distribution, successive sets of candidate solutions are pipelined through the network, from source to sinks and back. A time lag is incurred as the selected candidate travels through the network to undergo variation and fitness evaluation before it is inserted back into the population. This creates an age gap between the population from which a candidate solution is selected and the population into which it is inserted and leads to the question of how \emph{selection} and \emph{replacement} in the doubly distributed GA should proceed. The approach that is least efficient in terms of time, treats multiple pipelined sets of candidates as components of a single population that proceeds in an age-synchronized, generational style for selection and replacement. It sends pipelined sets of selected candidates through the network but waits until every set has emerged back out before replacing any of them. We show that this approach, which we term ``Generational/Single Population,'' incurs a cost of priming and flushing the pipeline but is faster than not pipelining at all.

To avoid intermittently flushing the pipeline and then needing to prime it again, our first approach is to divide the population into a number of subpopulations and insert selected then genetically varied individuals back into the same subpopulation they were selected from. Migration between sub-populations occurs at some specified frequency regardless of a slight age difference which maintains close temporal consistency. We call this approach ``Generational/Multi-population.''

Alternatively, we can be intentionally ``sloppy'' and forgo any temporal consistency. Much like a steady state GA, a single population is steadily updated. However, unlike a steady state GA, regardless of the time gap (between when a candidate is selected, genetically varied, then evaluated for fitness and when an attempt is made to insert it into the population), insertion simply proceeds with the current population as new candidates emerge processed from the network. In addition to yielding a simple algorithm, the ``temporally sloppy'' approach crudely approximates the asynchronously timed selection, reproduction and replacement events of a naturally evolving population. We dub this ``Non-generational/Single population.''

Pipelining increases the number of evaluations per time unit. The Generational/Multi-population and Generational/ Single population approaches are constrained to respect age synchrony between selection and replacement. But the Non-generational/Single population approach does not and, therefore, will have different and as yet unexplored dynamics. Will it converge with more or less fitness evaluations? Does the efficiency of pipelining produce a faster time to convergence? Will it find quality solutions? We explore these questions in the experiments.

Though the proposed algorithm is discussed in the context of network coding, the contributions of this paper are not limited within that scope. 1) A genetic algorithm with the proposed two novel methods of distribution can be readily applied to a variety of other optimization scenarios arising in communication networks (e.g., routing, resource allocation, etc.) or other connected systems where local decision variables are to be specified for the optimal performance of the whole system. 2) Furthermore, the proposed framework of temporal axis distribution can be combined with, not only the pipelining methods considered in this paper, a fairly general class of state-of-the-art strategies for parallel management of populations and communication between populations (e.g., \cite{ALN03, TM06}), because it imposes essentially no constraint on the implementation of any such strategies except that there is slight temporal inconsistency between populations, which as shown in this paper may also have little effect to other strategies.

The rest of the paper is organized as follows. Section~\ref{sec:nwcoding} describes and formulates the network coding problem. Section~\ref{sec:NCGA} describes the NCGA which serves as a baseline. Section~\ref{sec:DistHorizontal} motivates and describes distributing the NCGA along the \horizontal{} axis. Section~\ref{sec:DistVertical} motivates the distribution along the \vertical{} axis and describes three pipelined approaches. Section~\ref{sec:exp} experimentally quantifies the advantage of distribution on the \horizontal{} axis and compares the pipelined approaches. Section \ref{sec:con} concludes.

\section{Network Coding}\label{sec:nwcoding}

\begin{figure}[h]
\vspace{-0.2in}
\centerline{
\subfigure[Network $B$]{\includegraphics[height=1.2in]{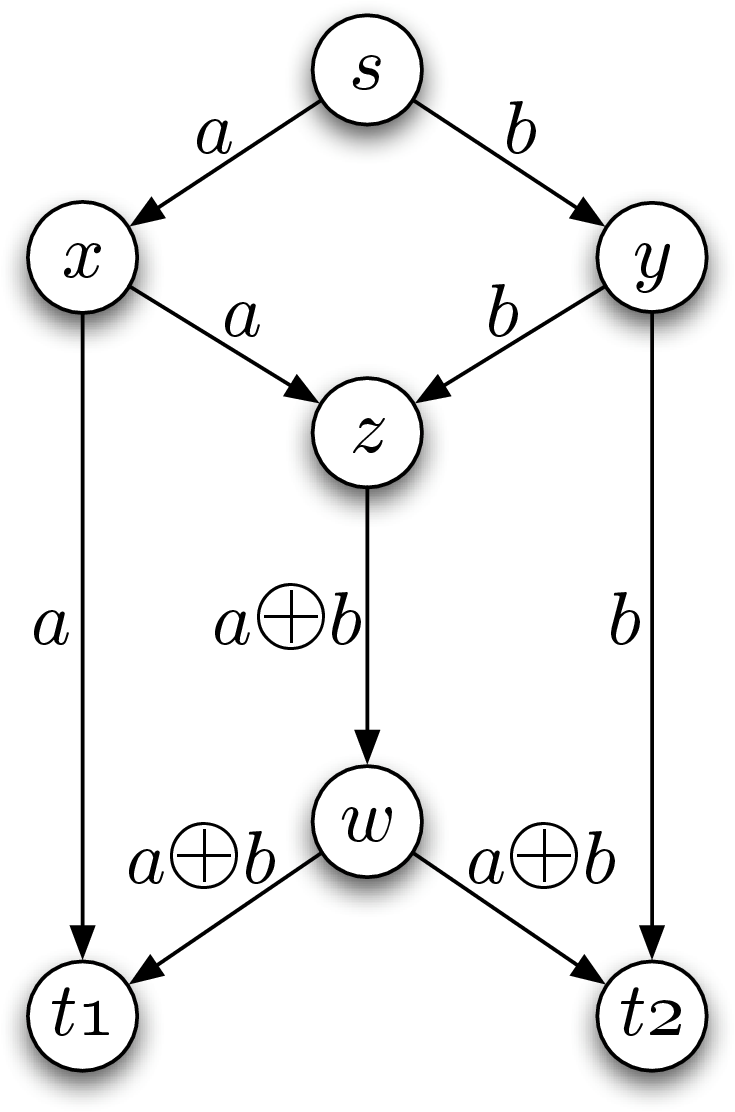}
\label{fig:bf}}
\hfil
\subfigure[Network $B'$]{\includegraphics[height=1.2in]{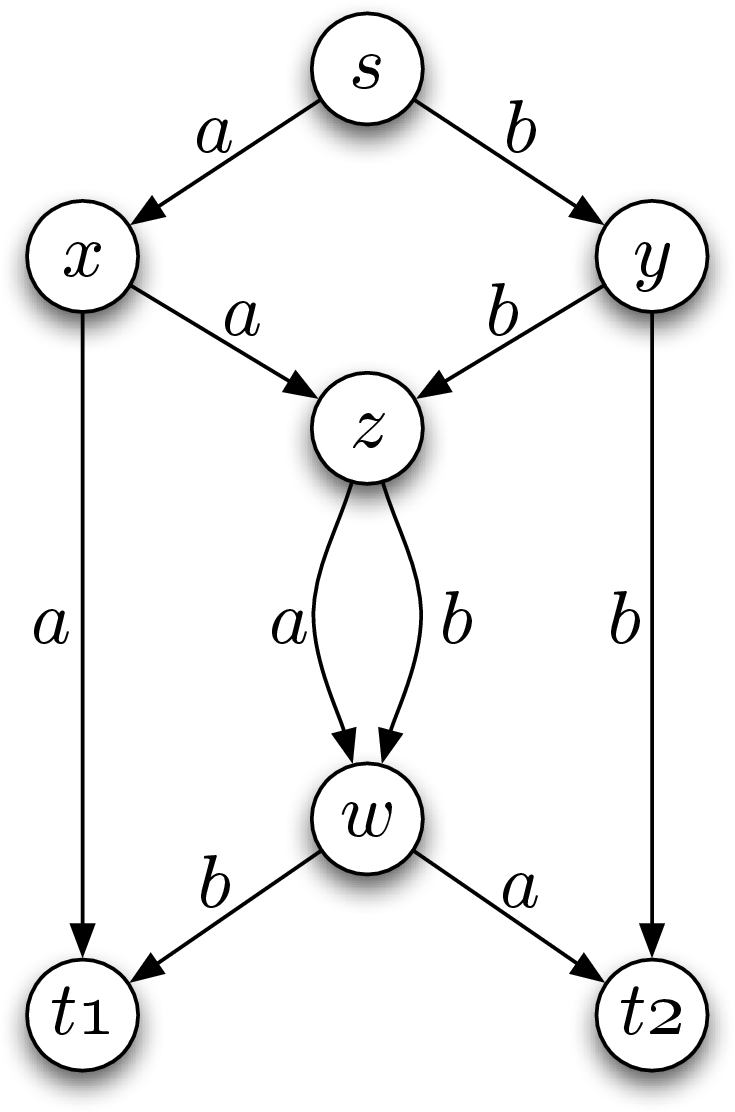}
\label{fig:bf2}}
\hfil
\subfigure[Network $C$]{\includegraphics[height=1.2in]{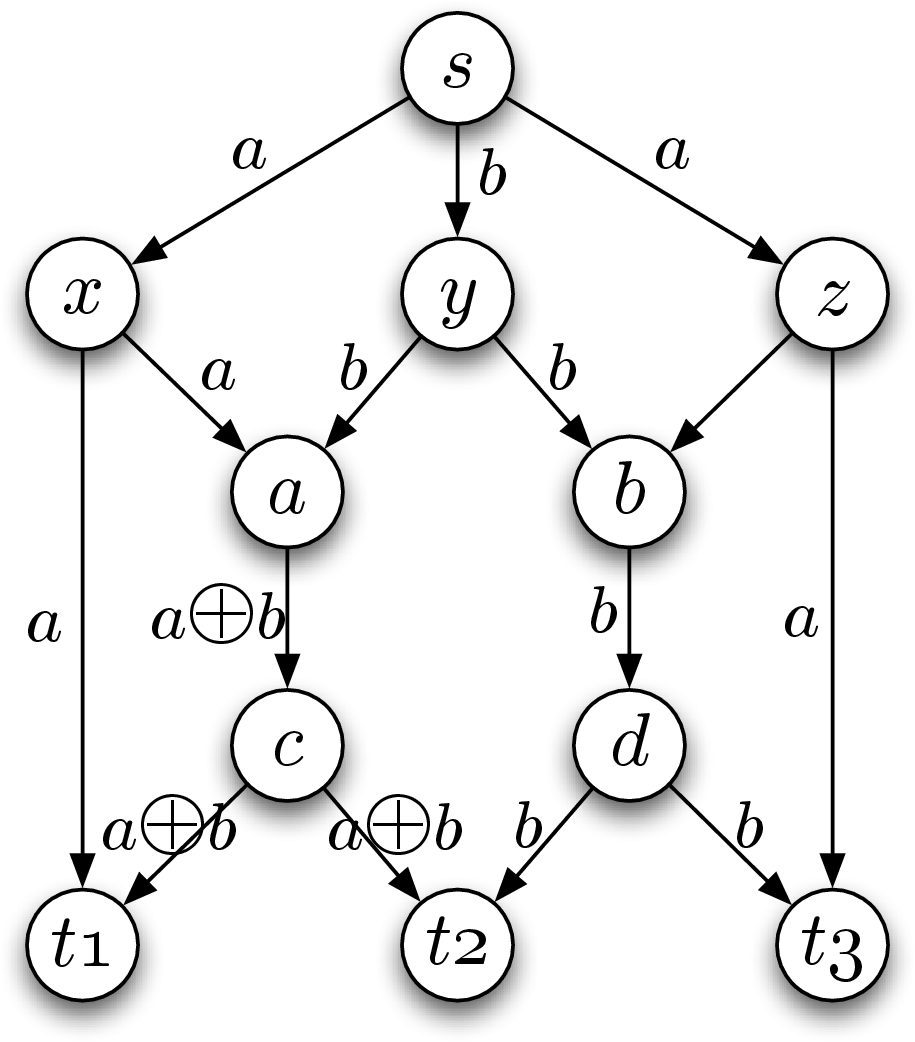}
\label{fig:corr}}
}
\vspace{-0.1in}
\caption{Sample networks for Example \ref{ex:intro}}
\vspace{-0.05in}
\label{fig:sample}
\end{figure}

Network coding is a novel technique that generalizes routing. In traditional routing, each interior network node, which is not a source or sink node, simply forwards the received data or sends out multiple copies of it. In contrast, network coding allows interior network nodes to perform arbitrary mathematical operations, e.g., summation or subtraction, to combine the data received from different links. It is well known that network throughput can be significantly increased by network coding \cite{ACLY00, LYC03}. While network coding is assumed to be done at all possible nodes in most of the network coding literature, it is often the case that network coding is required only at a subset of nodes to achieve the desired throughput. Consider Example 1:

\begin{ex}
In the canonical example of network $B$ (Figure \ref{fig:bf}) \cite{ACLY00}, where each link has unit capacity, source $s$ can send 2 units of data simultaneously to sinks $t_1$ and $t_2$, which is not possible with routing alone. But only node $z$ needs to combine its two inputs while all other nodes perform routing only. If we suppose that link $(z,w)$ in network $B$ has capacity 2, which we represent by two parallel unit-capacity links in network $B'$ (Figure \ref{fig:bf2}), a multicast of rate 2 is possible without network coding. In network $C$ (Figure \ref{fig:corr}), where node $s$ is to transmit data at rate 2 to the 3 leaf nodes, network coding is required either at node $a$ or at node $b$, but not at both.
\hfill $\square$
\label{ex:intro}
\end{ex}

Example \ref{ex:intro} leads us to the following question: To achieve the desired throughput, at which nodes does network coding need to occur? The problem of determining a minimal set of nodes where coding is required is NP-hard; its decision problem, which decides whether the given multicast rate is achievable without coding, reduces to a multiple Steiner subgraph problem, which is NP-hard~\cite{RP86}. For a GA, the problem can be posed as the minimization of coding cost (in links or nodes) subject to the constraint of feasibility (achieving the desired throughput).



\subsection{Problem Formulation}\label{sec:problem}

We assume that a network is given by a directed multigraph $G=(V,E)$ as in \cite{KM03} where each link has a unit capacity whose unit can be arbitrarily chosen, e.g., $P$ bits per second for a constant $P$, or a fixed size packet per unit time, etc. Links with larger capacities are represented by multiple links. Only integer flows are allowed, hence there is either no flow or a unit rate of flow on each link. We consider the single multicast scenario in which a single source $s \in V$ wishes to transmit data at rate $R$ to a set $T \subset V$ of sink nodes. Rate $R$ is said to be achievable if there exists a transmission scheme that enables all $|T|$ sinks to receive all of the information sent. We only consider linear coding, where a node's output on an outgoing link is a linear combination of the inputs from its incoming links. Linear coding is sufficient for multicast \cite{LYC03}.

Given an achievable rate $R$, we wish to determine a minimal set of nodes where coding is required in order to achieve this rate. However, whether coding is necessary at a node is determined by whether coding is necessary at at least one of the node's outgoing links and thus, as pointed out also in \cite{LSB06}, the number of coding links is in fact a more accurate estimator of the amount of computation incurred by coding. We assume hereafter that our objective is to minimize the number of coding \emph{links} rather than \emph{nodes}. 


It is clear that no coding is required at a node with only a single input since these nodes have nothing to combine with \cite{KAME06}. For a node with multiple incoming links, which we refer to as a \emph{merging node}, if the linearly coded output to a particular outgoing link weights all but one incoming message by zero, effectively no coding occurs on that link; even if the only nonzero coefficient is not identity, there is another coding scheme that replaces the coefficient by identity \cite{LSB06}. Thus, to determine whether coding is necessary at an outgoing link of a merging node, we need to verify whether we can constrain the output of the link to depend on a single input without destroying the achievability of the given rate. As in network $C$ of Example \ref{ex:intro}, the necessity of coding at a link depends on which other links code and thus the problem of deciding where to perform network coding in general involves a selection out of exponentially many possible choices. Employing a GA-based search method efficiently addresses the large and exponentially scaling size of the space.

\section{Network Coding GA ("A")}\label{sec:NCGA}

In the network research community, \cite{KAME06} and \cite{KMAO07} have documented results that demonstrate the benefit of the NCGA over other existing approaches in terms of reducing the number of coding links or nodes and its applicability to a variety of generalized scenarios. In the GA community, \cite{KAOM07} has investigated two different genotype encodings\footnote{To minimize confusion, throughout the paper, the term ``encoding" refers to ``genotype encoding" only, while the term ``coding" means ``network coding."} and associated operators. Reference \cite{KAOM07}'s main finding is that the encoding and the genetic operators that respect the block structure of the problem, which will be detailed later, substantially outperforms those do not. It is also claimed that such superior performance is mainly due to the modularity enforced by the block-wise genetic operators.


We first describe the elements of the NCGA that uses a standard generation-based GA control loop with centralized operations. This centralized NCGA, which we refer to as ``\textbf{Algorithm A}," serves as a baseline approach in comparison with the distributed versions of the algorithm, which share the GA elements introduced in this section.

\subsection{Genotype Encoding}

\begin{figure}[h]
\vspace{-0.1in}
\centerline{
\subfigure[Merging node $v$]{\includegraphics[height=1.4in]{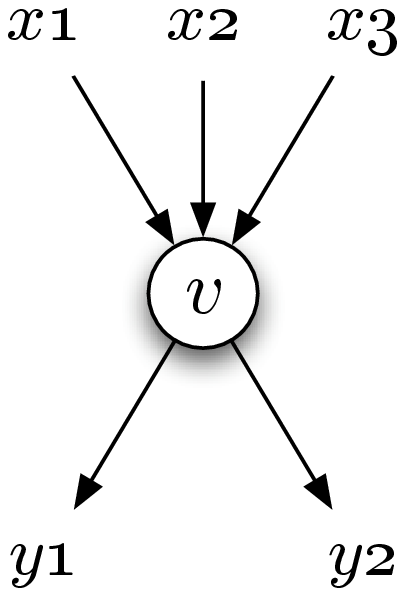}
\label{fig:block1}}
\subfigure[Two blocks for outgoing links of $v$]{\includegraphics[height=1.4in]{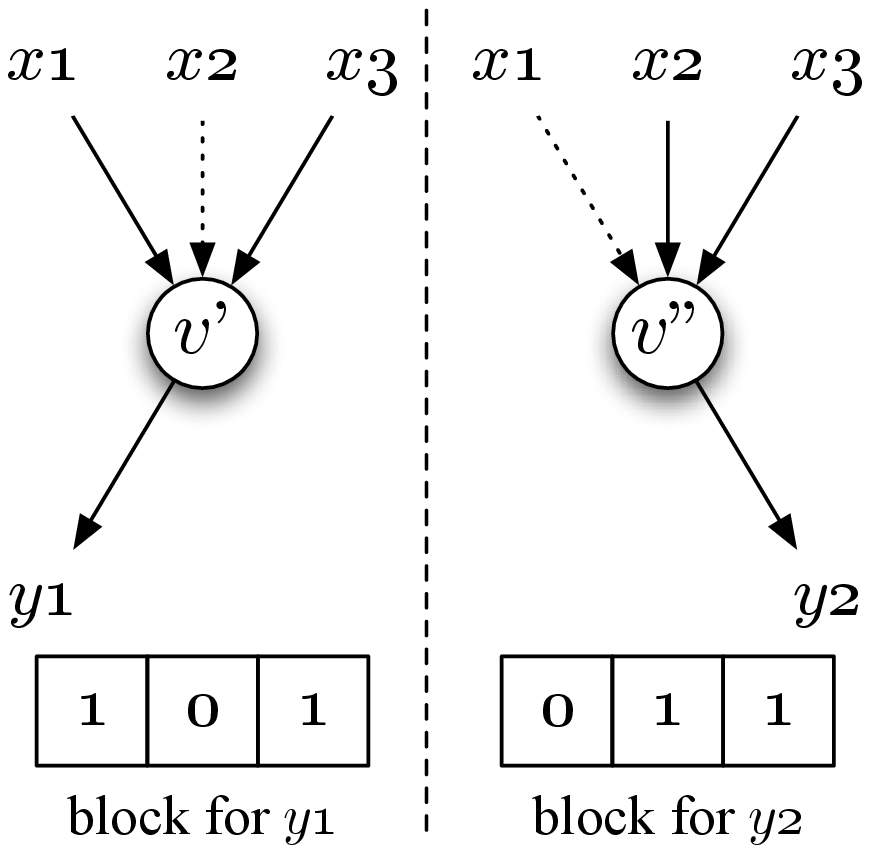}
\label{fig:block2}}
}
\vspace{-0.1in}
\caption{Node $v$ with 3 incoming and 2 outgoing links results in 2 blocks, each with 3 variables indicating the states of incoming links $(x_1, x_2, x_3)$ onto the associated outgoing link.}
\vspace{-0.05in}
\label{fig:block}
\end{figure}

Suppose a merging node with $k (\geq 2)$ incoming links. To consider the transmission to \emph{each} of its outgoing links, we assign a binary variable to each of its $k$ incoming links, whose being 1 indicates that the link state is \emph{active} (the input from the associated incoming link is transmitted to the outgoing link) and 0 indicates it is \emph{inactive}. Given that network coding is required for the transmission only if two or more link states are active, we may need to consider those $k$ variables together. We refer to the set of the $k$ variables as a \emph{block} of length $k$ (see Figure \ref{fig:block} for an example).

We notice that once a block has at least two 1's, coding is already required on the outgoing link associated with that block, and thus replacing all the remaining 0's with 1's has no effect on whether coding is done. Moreover, it can be shown that substituting 0 with 1, as opposed to substituting 1 with 0, does not hurt the feasibility. Therefore, for a feasible genotype (which is defined below), any block with two or more 1's can be treated the same as the block with all 1's. Thus we could group all the states with two or more active links into a single state, {\it coded} transmission. This state is rounded out by $k$ states for the {\it uncoded} transmissions of the input received from one of the $k$ single incoming links and one state indicating {\it no} transmission. Thus, each block of length $k$ can only take one of the following $(k+2)$ strings: $``111...1"$, $``100...0"$, $``010...0"$, $``001...0"$, $...$, $``000...1"$, $``000...0"$. If we denote by $d^v_{in}$ and $d^v_{out}$ the in-degree and the out-degree of node $v$, node $v$ has $d^v_{out}$ blocks of length $d^v_{in}$, and thus we have the search space of size $m=\prod_{v \in \mathcal{V}} (d^v_{in}+2)^{d^v_{out}}$, where $\mathcal{V}$ is the set of all merging nodes.

\subsection{Constraint and Fitness Function}

A genotype is called \emph{feasible} if there exists a network coding scheme that achieves the given rate $R$ with the link states determined by the genotype. For the feasibility test of a genotype, we rely on the algebraic method described in \cite{KMAO07}, which later enables a distributed feasibility test. Given the feasibility of genotype $\underline{y}$, its fitness value $F$ is assigned as
\begin{equation*}
F(\underline{y})=
    \begin{cases}
        \text{number of coding links}, & \text{if } \underline{y} \text{ is feasible}, \\
        \infty, & \text{if } \underline{y} \text{ is infeasible},
    \end{cases}
\end{equation*}
where the number of coding links can be easily calculated by counting the number of blocks in the genotype with at least two $1$'s.

\subsection{Genetic Operators}

To preserve the above encoding structure, we need to define a new set of genetic operators, which we refer to as \emph{block-wise} genetic operators. For block-wise uniform crossover, we let two genotypes subject to crossover exchange each block, rather than bit, independently with the given crossover probability. For block-wise mutation, we let each block under mutation take another string chosen uniformly at random out of $(k+1)$ other strings for a length-$k$ block.

\subsection{Other Elements}

The NCGA evaluates fitness in a multi-step way: 1) each merging node consults the corresponding genotype blocks to compute \emph{random} linear combinations of the inputs\footnote{See \cite{HKM03} for an explanation of why this is sufficient.}, 2) alternately routed messages reach the sinks, 3) the feasibility of the genotype is assessed at the sinks, 4) if feasible, the coding links are counted.

The NCGA uses tournament selection and terminates at some maximum number of generations. Afterward, the best solution of the run is optimized with \emph{greedy sweep}: each of the remaining 1's is switched to 0 if it can be done without violating feasibility. This procedure can only improve the solution, and sometimes the improvement can be substantial \cite{KMAO07}. 


\section{\HorizontalTitle{} Axis Distribution ("B")}\label{sec:DistHorizontal}

Decentralizing the NCGA enables a network coding protocol where the resources used for coding are optimized on the fly in a setup phase. Plus, distribution reduces the computational efficiency of the algebraic feasibility test (see Section \ref{sec:Complexity} for details). We refer to this \horizontal{}(only)-distributed NCGA as ``\textbf{Algorithm B}."

\subsection{Overview}\label{sec:genoOverview}
\begin{figure}[h]
\vspace{-0.05in}
    \centering
    \includegraphics[height=1.6in]{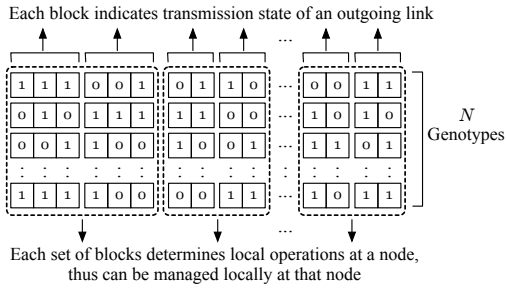}
    \vspace{-0.1in}
    \caption{Structure of Population}
    \label{fig:PopStructure}
\vspace{-0.15in}
\end{figure}

\begin{figure}[h]
    \centering
    \includegraphics[height=1.8in]{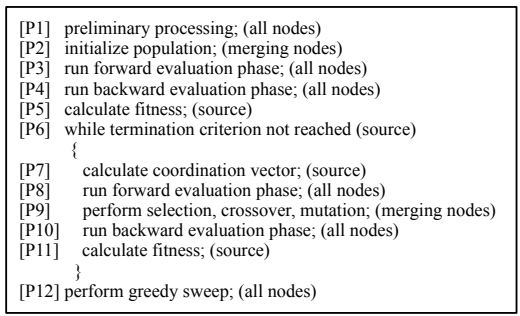}
    \vspace{-0.1in}
    \caption{Flow of \Horizontal{}-Distributed NCGA}
    \label{fig:distributed}
\vspace{-0.05in}
\end{figure}

Because of the way network coding depends on each merging node contributing to the coding, and because each merging node references its corresponding block on a genotype, the appropriate way to distribute the NCGA is to have each node handle only the blocks it needs from every member of the population. So, instead of dividing up the population and giving each island a subset of genotypes, we divide up the genotype of every population member and give each merging node a population wide set of that genotype subset. Thus, in contrast to a conventional distributed GA, the axis of distribution is \horizontal{} rather than population as illustrated in Figure \ref{fig:PopStructure}. The previously centralized fitness evaluation steps are transformed into: 1) forward evaluation stage from merging nodes to each sink 2) backward evaluation stage from sinks to source and 3) fitness calculation at the source. With some amount of additional message information and coordination, all genetic operations can be done locally at each merging node. See Figure~\ref{fig:distributed} for the overall structure.

\subsection{Assumptions}
\label{sec:Assumptions}

While we assume that each link can transmit one packet with the fixed size, say $P$ bits, per time unit in the given direction, each link is also assumed to be able to send some amount of feedback data, typically much smaller than the packet size, in the \emph{reverse} direction. Also, we assume that each interior node operates in a burst-oriented mode; i.e., for the forward (backward) evaluation phase, each node starts updating its output only after an updated input has been received from all incoming (outgoing) links.


\subsection{Details of \Horizontal{}-Distributed Algorithm}

\subsubsection{Preliminary Processing \emph{[P1]}}

The source initiates the algorithm by transmitting the ``optimize" signal containing the following predetermined parameters: target multicast rate $R$, population size $N$, the size $q$ of the finite field to be used, crossover probability, and mutation rate. Each participating node that has received the signal passes the signal to its downstream nodes.

\subsubsection{Population Initialization \emph{[P2]}}


Each merging node with $d_{in} (\geq 2)$ incoming links will manage a \emph{coding vector} indicating the link states per population member. To initialize its subset of the population, each merging node generates $N \cdot d_{in} \cdot d_{out}$ binary numbers randomly. Then, for the coding vectors corresponding to the first of the $N$ chromosomes, all the components are set to 1 \cite{KAME06}.

\subsubsection{Forward Evaluation Phase \emph{[P3, P8]}}

For the feasibility test of a chromosome, each node transmits a vector consisting of $R$ components, which we refer to as a \emph{pilot vector}. Each of its the components is from the finite field $\mathbb{F}_q$ and the $i$-th component represents the coefficient used to encode the $i$-th source data. We assume that a set of $N$ pilot vectors is transmitted together by a single packet.

The source initiates the forward evaluation phase by sending out on each of its outgoing links a set of $N$ random pilot vectors. Each non-merging node simply forwards all the pilot vectors received from its incoming link to all its outgoing links.

Each merging node transmits on each of its outgoing links a random linear combination of the received pilot vectors, computed based on the node's coding vectors as follows. Let us consider a particular outgoing link and denote the associated $d_{in}$ coding vectors by $v_1$, $v_2$, ..., $v_{d_{in}}$. For the $i$-th ($1 \leq i \leq N$) output pilot vector $u_i$, we denote the $i$-th input pilot vectors received form the incoming links by $w_1$, $w_2$, ..., $w_{d_{in}}$. Define the set $J$ of indices as
\begin{equation*}
J = \{ 1 \leq j \leq d_{in} | \text{ the $i$-th component of $v_j$ is 1}\}.
\end{equation*}
Then,
\begin{equation*}
u_i = \sum_{j \in J} w_j \cdot \text{rand} (\mathbb{F}_q),
\end{equation*}
where rand$(\mathbb{F}_q)$ denotes a random element from $\mathbb{F}_q$. If the set $J$ is empty, $u_i$ is assumed to be zero.

\subsubsection{Backward Evaluation Phase \emph{[P4, P10]}}

To calculate a chromosome's fitness value, two kinds of information need to be gathered: 1) whether each sink can decode data of rate $R$ and 2) how many links are used for coding at each merging node. 

Each sink can determine whether data of rate $R$ is decodable for each of the $N$ chromosomes by computing the rank of the collection of received pilot vectors. It is worth to point out that this is the same algebraic evaluation method described in \cite{KAME06}, but the difference is that, rather than computing the system matrix with randomized elements centrally, now we actually construct random linear codes over the network in a decentralized fashion. Hence, this feasibility test also bears the same, but uncritical, possibility of errors as in the centralized case. Regarding the number of coding links, each merging node can simply count the number links where coding is required by inspecting its coding vectors used in the forward evaluation phase.

For the feedback of this information, each node transmits a vector consisting of $N$ components, which is referred to as a \emph{fitness vector}. The backward evaluation phase proceeds as follows:


\renewcommand{\labelenumi}{$\bullet$}
\begin{enumerate}
\item After the feasibility tests of the $N$ chromosomes are done, each sink generates a fitness vector whose $i$-th ($1 \leq i \leq N$) component is zero if the $i$-th chromosome is feasible at the sink, and infinity otherwise. Each sink then initiates the backward evaluation phase by transmitting its fitness vector to all of its parents.
\item Each interior node calculates its own fitness vector whose $i$-th ($1 \leq i \leq N$) component is the number of coding links at the node for the $i$-th chromosome plus the sum of all the $i$-th components of the received fitness vectors. Each node then transmits the calculated fitness vector to \emph{only one} of its parents, and an all-zero fitness vector (for just signaling) to the other parent nodes.
\end{enumerate}
Note that, since the network is assumed to be acyclic, each coding link of a chromosome contributes exactly once to the corresponding component of the source node's fitness vector, and thus the above update procedure provides the source with the correct total number of coding links.

\subsubsection{Fitness Calculation \emph{[P5, P11]}}

The source calculates the fitness values of $N$ chromosomes simply by component-wise summation of the received fitness vectors. Note that if an infinity were generated by \emph{any} of the sinks, it should dominate the summations all the way up to the source, and thus the source can calculate the correct fitness value for the infeasible chromosome.

\subsubsection{Termination Criterion \emph{[P6]}}

The source can determine when to terminate the optimization by counting the number of generations iterated thus far.

\subsubsection{Coordination Vector Calculation \emph{[P7]}}

Since the population is divided into subsets that are managed at the merging nodes, genetic operations also need to be done locally at the merging nodes. However, some amount of coordination is required for consistent genetic operations throughout all the merging nodes, more specifically, for 1) selection of chromosomes, 2) paring of chromosomes for crossover, and 3) whether each pair is subject to crossover. This information is carried by a \emph{coordination vector}, calculated at the source, consisting of the indices of selected chromosomes that are randomly paired and 1-bit data for each pair indicating whether the pair needs to be crossed over. The coordination vector is \emph{transmitted together with} the pilot vectors in the next forward evaluation phase.

\subsubsection{Genetic Operations \emph{[P8]}}

Based on the received coordination vector, each merging node can locally perform genetic operations and renew its portion of the population as follows:
\renewcommand{\labelenumi}{$\bullet$}
\begin{enumerate}
\item For selection, each node only retains the coding vectors that correspond to the indices of selected chromosomes.

\item For block-wise crossover, each node independently determines whether each block is crossed over. Since no block is shared by multiple merging nodes, this can be done independently at each merging node.

\item For block-wise mutation, each node independently determines whether each block is mutated without any coordination with other nodes either.
\end{enumerate}

\subsubsection{Greedy Sweep \emph{[P12]}}

Greedy sweep requires an additional protocol where, after the iteration terminates, the source is notified of the merging nodes with at least one coding link, for each of which the source sends out a packet to test if uncoded transmission is possible on the link(s) where currently coding is required. Since this additional protocol requires more extensive coordination between nodes, we may leave this procedure optional, whose detailed description is omitted owing to space limitations. 

\subsection{Complexity}\label{sec:Complexity}


The computational complexity required for evaluation of a single chromosome is $O(\sum_{v \in \mathcal{V}} d^v_{in}d^v_{out}R + \sum_{w \in V \setminus \mathcal{V}} d^w_{out} + \sum_{t \in T} {d^t_{in}}^2 R)$, which can be substantially less than that for the centralized version of the algorithm, i.e., $O(|T|\cdot(|E|^{2.376}+R^3))$ or $O(|T|\cdot(|E'|^2 \sqrt{|V'|}))$ \cite{KMAO07}.

\begin{figure*}[t]
\centering
\subfigure[Timing Diagram of Algorithm B: Genotype(only)-distributed NCGA.]{\includegraphics[height=1.1in]{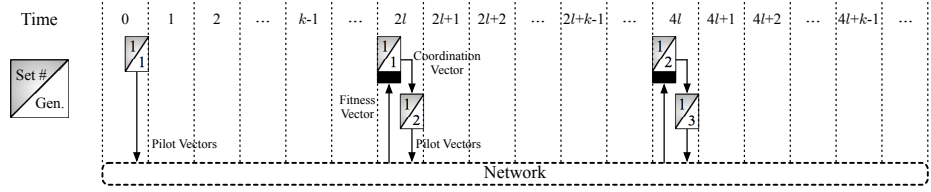}
\label{fig:Baseline}}
\\
\subfigure[Timing Diagram of Algorithm D: This doubly distributed algorithm (Generational/ Multi-population) exploits pipelining, does not require intermittent flushing, respects age consistency between selected and replecement, and respects close age consistency in migration.]{\includegraphics[height=1.1in]{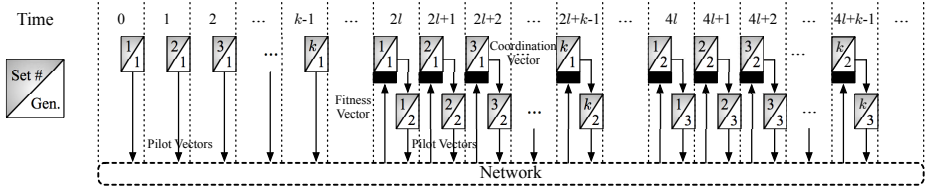}
\label{fig:Parallel}}
\vspace{-0.1in}
\caption{Comparison of Algorithms B and D via Timing Diagrams. }
\vspace{-0.1in}
\end{figure*}

\section{\VerticalTitle{} Axis Distribution}\label{sec:DistVertical}

A unique characteristic of the \horizontal{}-distributed NCGA is that once each generation is initiated at the source (procedure [P7] in Figure \ref{fig:distributed}), the fitness values of $N$ genotypes become only available after the forward and backward evaluation phases are done, i.e., when the last fitness vector arrives at the source. Let us assume that the time required for each node to calculate its outgoing pilot vectors based on the received ones is negligible compared with the time required for packet transmissions. Then, if we denote by $l$ the length of the longest path from the source to any of the sinks, the time lag between the initiation of the generation and the termination of the backward evaluation phase is $2l$ time units (see Figure \ref{fig:Baseline}).

Let us now define the \emph{evaluation efficiency}, which we denote by $\varepsilon_v$, as the number of fitness evaluations performed per unit time throughout the iteration of the GA. Then, for Algorithm B(\horizontal{}(only)-distributed NCGA), $\varepsilon_v$ is only $N/2l$.

For better efficiency, we may still utilize the network resources, while waiting for the fitness vectors to return to the source, to evaluate more genotypes. Suppose that, after initiating the forward evaluation phase of the $n$-th generation at time $t$, we initiate additional $k-1$ forward evaluation phases at times $t+1$, ..., $t+k-1$. When $k=2l$, the network resources become fully utilized by the time when the fitness values of the first set of $N$ genotypes are available. Note that in fact $k$ may even exceed $2l$, but then the evaluation of the $(n+1)$-th generation starts delayed at time $t+k$, rather than $t+2l$. For simplicity, we assume $k \leq 2l$ in the following.

\subsection{Generational / Single Population ("C")}

If we consider the $k$ sets of $N$ genotypes as a single population, we have to wait additional $k-1$ time units, after the first backward evaluation phase ends (at time $t+2l$), to proceed to the next generation. In other words, we must flush the pipeline (and prime it again). Hence, the evaluation efficiency is given by
\begin{equation*}
\varepsilon_v=\frac{kN}{2l+k-1},
\end{equation*}
whose maximum is obtained when $k=2l$ such that $\varepsilon_v=\frac{2lN}{4l-1} \approx \frac{N}{2}$. For later comparison, we refer to this algorithm with $k=2l$ as ``\textbf{Algorithm C}."

Avoiding the inefficiency of flushing the pipeline would generate a better $\varepsilon_v$ and consequently faster convergence, provided that the algorithm requires a similar number of evaluations for the solutions of the same quality. Depending on how to manage those $k$ sets of $N$ genotypes, we may consider two different approaches as follows.



\subsection{Generational / Multi-Population ("D")}

In this approach, referred to as ``\textbf{Algorithm D}," we regard each of those $k$ sets of $N$ genotypes as a \emph{subpopulation} which occasionally exchanges individuals with other subpopulations. It is worth to point out that, unlike typical island parallel GAs \cite{CP98} where subpopulations are \emph{spatially} distributed over different locations of computation, we have subpopulations that are \emph{temporally} distributed over different times of evaluation.

We assume that migration is done at every $f$ generations such that, before selection, each subpopulation replaces its worst $k-1$ individuals with the collection of $k-1$ individuals, one from each of the other $k-1$ subpopulations. Since we have no constraint on the (spatial) connections between the subpopulations, we can freely choose to assume and exploit the complete connectivity between subpopulations.

On the other hand, our algorithm imposes a different kind of constraint on migration, which is regarding the time synchronization between subpopulations. Let us assume that there is no delay in the network, so the backward evaluation phase of a particular subpopulation ends exactly after $2l$ time units its forward evaluation phase started. Suppose now that migration is about to happen at time $t+1$ while constructing the first subpopulation for the $(n+1)$-th generation. At that time, only the first subpopulation has the fitness values for the $n$-th generation, while all other $k-1$ subpopulations still wait for their fitness values for the $n$-th generation to become available. Similarly, at time $t+j$ $(1 \leq j \leq k)$, only the first $j$ subpopulations have their fitness values for the $n$-th generation, while the remaining $k-j$ subpopulations do not. If we choose to perform migration in a \emph{age-synchronized}, i.e., \emph{temporally consistent} manner such that all the subpopulations exchange the best individuals of the \emph{same} generation, we have to wait until time $t+k$ without being able to renew any subpopulation. Hence, we alternatively perform the \emph{age-mixed}, i.e., \emph{temporally closely consistent}, migration, where we collect the best individuals from the other $k-1$ subpopulations of the \emph{most recent} generation for which the fitness values are available. For instance, when we renew the $j$-th $(2 \leq j \leq k-1)$ subpopulation at time $t+j$, we take the best individual from each of the $1, ..., (j-1)$-th subpopulations at generation $n$, and from each of the $(j+1), ..., k$-th subpopulations at generation $n-1$.

Algorithm D proceeds in a completely pipelined manner (see Figure \ref{fig:Parallel}), yielding the evaluation efficiency
\begin{equation*}
\varepsilon_v=\frac{gkN}{(g+1)2l+k-1},
\end{equation*}
where $g$ is the number of generations at the termination of the iteration. Note that, when $k=2l$ and $g \gg 1$, $\varepsilon_v \approx N$.

Note that most changes in Algorithm D, compared with the \horizontal{}-distributed NCGA in Section \ref{sec:DistHorizontal}, are regarding the computational aspects at the source. Hence, Algorithm D can be implemented within the same framework the \horizontal{}-distributed NCGA, with slight changes in the structure of the coordination vector and the increased number of coding vectors that each merging node keeps. Owing to space limits, further implementational details are omitted.

\subsection{Non-Generational / Single Population ("E")}

\begin{figure*}[t]
    \centering
    \includegraphics[height=1.1in]{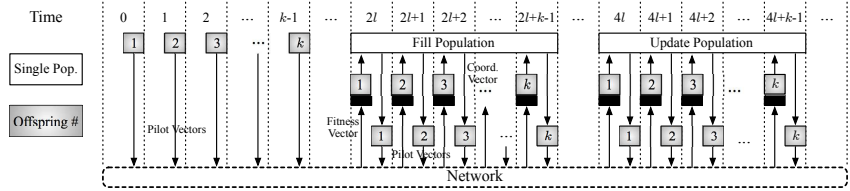}
    \vspace{-0.1in}
    \caption{Timing Diagram of Algorithm E: This doubly distributed algorithm (Non-generational/Single population) exploits pipelining and  does not require intermittent flushing. It is ``sloppy'' with respect to temporal consistency between selection and replacement by using a single population with just-in-time updating.}
    \label{fig:SS}
\vspace{-0.1in}
\end{figure*}

Rather than managing $k$ separate subpopulations, this approach, referred to as ``\textbf{Algorithm E}," operates on a single population of size $M=kN$. The population is updated when the fitness values of each of the $k$ sets of $N$ genotypes, referred to as \emph{offspring}, become available (i.e., ``just-in-time''). This is a temporally ``sloppy`` approach. From time 1 to $k$, the forward evaluation phases for the initial (random) $k$ offspring are initiated. At time $2l+j$ $(1 \leq j \leq k)$, the fitness values for the $j$-th offspring can be calculated at the source and \emph{all} those $N$ genotypes are just added to the population. We then calculate the coordination vector for the $j$-th offspring, by performing tournament selection out of the current population, which is partially filled until time $2l+k$, and initiate the forward evaluation phase for the second generation. At time $4l+j$ $(1 \leq j \leq k)$ and on, we update the population as follows: First combine the $j$-th offspring, whose fitness values are just calculated, with the existing population, and then pick the best $kN$ individuals, out of those $(k+1)N$ individuals, to form the updated population.
 
Considering each window of $2l$ time units from the beginning, we notice that except for the first and the last windows, $kN$ genotypes are evaluated in each window (see Figure \ref{fig:SS}). Hence, if we assume that the total number of elapsed time units is large ($\gg 1$), we have $\varepsilon_v \approx \frac{kN}{2l}$, and when $k=2l$, we obtain the maximum $\varepsilon_v \approx N$.

Algorithm E can also be implemented similarly to the \horizontal{}-distributed NCGA with some changes in the coordination and coding vectors, whose details are omitted.


\section{Experiments} \label{sec:exp}

\subsection{Effect of \Horizontal{} Axis Distribution}

Since the \horizontal{}-distributed NCGA (Algorithm B) shares the same computational part of GA with the centralized one (Algorithm A), the two algorithms show the same performance in terms of solution quality. However, as described in Section \ref{sec:Complexity}, the computational complexity required by Algorithm B depends only on local topological parameters, which can often lead to a significant gain in terms of the running time. To compare the elapsed running time of the two algorithms, we run a test on a created set of topologies with high connectivity such that there exists a link between each pair of numbered nodes $i$ and $j$ ($i<j$), where the source is node 1 and the sinks are the last 10 nodes. 
The test is done by a simulation on a single machine while each node's function is performed by a separate thread, thus it is pessimistic since it cannot benefit from the multi-processing gain whereas it only suffers from additional computational burdens for managing a number of threads. Table \ref{tab:Eval3} shows that, nevertheless, Algorithm B exhibits an advantage in running time as the size of the network grows.

\begin{table}[h]
\centering{
\begin{tabular}{|c|c|c|c|c|c|c|} \hline
    Number of nodes & 15 & 20 & 25 & 30 & 35 & 40 \\ \hline
    Algorithm A & 0.3 & 1.5 & 4.3 & 13.5 & 29.5 & 65.6 \\ \hline
    Algorithm B & 1.8 & 2.7 & 4.4 & 6.3 & 10.8 & 15.4 \\ \hline
\end{tabular}
}
\vspace{-0.1in}
\caption{Running Time Per Generation (seconds)}
\label{tab:Eval3}
\vspace{-0.1in}
\end{table}

\subsection{Effect of \Vertical{} Axis Distribution}

To compare the doubly distributed approaches, we construct network $G$ by cascading 15 copies of network $B'$ in Example \ref{ex:intro}(Figure \ref{fig:bf2}) in the form of a depth-4 binary tree such that the source of each subsequent copy of $B'$ is replaced by an earlier copy's sink. The source is the tree's root node and the sinks are the 16 leaf nodes. Setting $P$, the unit packet size, to 1500 bytes as a typical ethernet packet, we can calculate that $N$, the number of genotypes handled by a single packet, is around 200. Since $l=16$ in network $G$, $k=2l=32$.



\begin{table}[h]
\centering{
\begin{tabular}{|c|l|}\hline
 & \multicolumn{1}{c|}{Parameters on Population} \\ \hline
B & Pop. size: 200 \\ \hline
C & Pop. size: 6400 \\ \hline
\multicolumn{1}{|l|}{D$_{10}$} & Subpops. (size, \#): (200,32), Migration freq.: 10 \\ \hline
\multicolumn{1}{|l|}{D$_{1}$} & Subpops. (size, \#): (200,32), Migration freq.: 1 \\ \hline
E & Pop. size: 6400, Offspring size: 200 \\ \hline
\end{tabular}
}
\vspace{-0.1in}
\caption{Population Parameters for Algorithms}
\label{tab:PopParams}
\vspace{-0.05in}
\end{table}

Table \ref{tab:PopParams} summarizes the parameters for five algorithms we experiment with. Migration frequency ($f$) is changed from 10 to 1 from Algorithm D$_{10}$ to D$_{1}$. We set the tournament size to the half of the (sub)population size in each algorithm, i.e., 100, 3200, 100, 3200 for Algorithms B, C, D, E, respectively. The mixing ratio and the crossover probability are both 0.8 and the mutation rate is 0.015 for all algorithms. We perform 30 runs for each algorithm until the algorithm converges to the optimal solution, which for network $G$ is known to be zero. Table \ref{tab:Result} shows the elapsed time units with the \emph{time efficiency} $\varepsilon_t$, which we define as the algorithm's speedup with respect to Algorithm B, and the total number of evaluations with the evaluation efficiency $\varepsilon_v$ obtained from the experiments, which indeed matches the theoretical values almost exactly. For elapsed time and number of evaluations, p-value resulting from paired t-test with the next best (i,e., smallest) one is reported.

\begin{table}[h]
\centering{
\begin{tabular}{|c|r|c|c|r|c|r|}\hline
 & \multicolumn{1}{c|}{Time} & p-value & $\varepsilon_t$ & \multicolumn{1}{c|}{\#Eval} & p-value & \multicolumn{1}{c|}{$\varepsilon_v$} \\ \hline
B & 13,907 & - & 1.00 & 86,920 & 1.38e-14 & 6.25 \\ \hline
C & 5,427 & 1.66e-08 & 2.56 & 542,720 & 2.10e-03 & 100.00 \\ \hline
\multicolumn{1}{|l|}{D$_{10}$} & 2,497 & 1.58e-04 & 5.57 & 492,920 & 0.307 & 197.44 \\ \hline
\multicolumn{1}{|l|}{D$_{1}$} & 4,157 & 7.55e-03 & 3.35 & 824,980 & - & 198.46 \\ \hline
E & 3,968 & 0.691 & 3.50 & 781,100 & 0.691 & 198.39 \\ \hline
\end{tabular}
}
\vspace{-0.1in}
\caption{Result of Experiments}
\label{tab:Result}
\vspace{-0.05in}
\end{table}

\begin{figure}[h]
\vspace{-0.1in}
    \centering
    \includegraphics[height=1.8in]{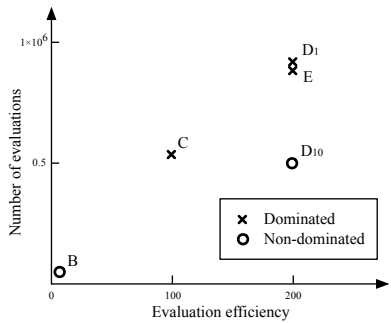}
    \vspace{-0.1in}
    \caption{Tradeoff Plot}
    \label{fig:tradeoff}
\vspace{-0.1in}
\end{figure}

Pipelining is intended to be efficient by reducing the idle time of network nodes, hence Algorithm B, which does not pipeline, has the lowest $\varepsilon_v$. Though Algorithm C, which pipelines but stop to flush and re-prime, has much increased $\varepsilon_v$, Algorithms D$_{10}$, D$_{1} $, and E, which operate fully pipelined, offer the highest $\varepsilon_v$. Note, however, that the different dynamics of these algorithms may impact the number of fitness evaluations required to reach the optimal solution, hence as can be observed in Table \ref{tab:Result}, the number of evaluations (and consequently, the realized $\varepsilon_t$) do not reveal $\varepsilon_v$ in proportion. Figure \ref{fig:tradeoff} shows that evaluation efficiency comes at the cost of additional fitness evaluations. Algorithms D$_{10}$ and B dominate all others yet not each other; Algorithm B is less efficient (it does not pipeline) but requires less fitness evaluations, while D$_{10}$ is more efficient but requires more evaluations. Algorithm D$_{10}$ gives a speedup ($\varepsilon_t$) of more than 5 times over algorithm B. Algorithms C, D$_{1}$ and E, though dominated by D$_{10}$, still offer higher $\varepsilon_t$ than B. These algorithms thus merit additional investigation because they may give better performance for different network topologies or other problems.


Algorithms D$_{1}$ and D$_{10}$, though distributed temporally, resemble a spatially distributed GA (referred to as multiple-deme GA in \cite{CP98}) in that they incur no communication overhead and can assume a fully-connected processor topology. The only difference in algorithm dynamics is that migration takes place between sub-populations that differ in age by one generation (see Section 5.2). Thus the performances of D$_{1}$ and D$_{10}$ as compared to B are in fact foreseeable from the observation that, in general, multiple-deme GAs require a greater number of evaluations than a standard GA while offering speedups due to parallelism, which is equivalent to higher $\varepsilon_v$. However, in our experiments, the size and the number of subpopulations are determined to maximize $\varepsilon_v$ rather than the performance of GA. Determining the migration strategy for multiple-GAs is an open question and probably problem dependent \cite{CPG00}.  

Algorithm E is a completely new algorithm, where the selection from the population and the replacement of offsprings are temporally inconsistent. A (slightly) similar property can be found in the second prototype for parallel GA in \cite{Gre81}, where the algorithm sends out individuals to processors to be evaluated, and inserts and re-selects them opportunistically, i.e., when their fitness becomes available. Such, rather radical, changes in algorithm dynamics may raise a question whether Algorithm E would even work, which is verified by our experiments. The performance of $E$ is similar to that of D$_{1}$, hence surpassed by D$_{10}$, which can be explained by the observation that the temporal mixing of E is similar to D$_{1}$'s frequent mixing. Together, these two results suggest that the doubly distributed GA is robust to age mixing (i.e., temporal sloppiness), which deserves further in-depth analysis in the future.

\section{Conclusions} \label{sec:con}

We have presented a GA which is distributed in two novel ways: along \horizontal{} and \vertical{} axes. In order to distribute the fitness evaluation for the network coding problem, our doubly distributed algorithm first distributes, for every member of the population, a subset of the genotype to each network node rather than a subset of the population to each. To maximize the efficient use of the computational nodes in the network, the second axis divides the candidate solutions into pipelined sets and thus the distribution is in the temporal domain, rather that in the spatial domain. We have found that this \vertical{} distribution may lead to temporal inconsistency in selection and replacement, however our experiments have yielded better efficiency in terms of the time to convergence without incurring significant penalties.


\begin{thebibliography}{50}
\vspace*{0.5mm}
\scriptsize


\bibitem{ACLY00}
R.~Ahlswede, N.~Cai, S.-Y.~R. Li, and R.~W. Yeung.
\newblock Network information flow.
\newblock {\em {IEEE} Trans. Inform. Theory}, 46(4):1204--1216, 2000.

\bibitem{ALN03}
E.~Alba, F.~Luna, and A.~J. Nebro.
\newblock Parallel heterogeneous genetic algorithms for continuous
  optimization.
\newblock In {\em Proc. IPDPS}, 2003.

\bibitem{CP98}
E.~Cant\'{u}-Paz.
\newblock A survey of parallel genetic algorithms.
\newblock {\em Calculateurs Parall\`{e}les, R\'{e}seaux et Syst\`{e}ms
  R\'{e}partis}, 10(2):141--171, 1998.

\bibitem{CPG00}
E.~Cant\'{u}-Paz and D.~E. Goldberg.
\newblock Efficient parallel genetic algorithms: {T}heory and practice.
\newblock {\em Comput. Methods Appl. Mech. Engrg.}, 186:211--238, 2000.

\bibitem{Gre81}
J.~J. Grefenstette.
\newblock Parallel adaptive algorithms for function optimization.
\newblock Technical Report CS-81-19, Vanderbilt Univ. Computer Science Dept.,
  1981.

\bibitem{HKM03}
T.~Ho, R.~Koetter, M.~M\'{e}dard, D.~R. Karger, and M.~Effros.
\newblock The benefits of coding over routing in a randomized setting.
\newblock In {\em Proc. {IEEE} ISIT}, 2003.

\bibitem{KAOM07}
M.~Kim, V.~Aggarwal, U.-M. O'Reilly, and M.~M\'{e}dard.
\newblock Genetic representations for evolutionary minimization of network
  coding resources.
\newblock In {\em Proc. EvoComnet}, 2007.

\bibitem{KAME06}
M.~Kim, C.~W. Ahn, M.~M\'{e}dard, and M.~Effros.
\newblock On minimizing network coding resources: An evolutionary approach.
\newblock In {\em Proc. NetCod}, 2006.

\bibitem{KMAO07}
M.~Kim, M.~M\'{e}dard, V.~Aggarwal, U.-M. O'Reilly, W.~Kim, C.~W. Ahn, and
  M.~Effros.
\newblock Evolutionary approaches to minimizing network coding resources.
\newblock In {\em Proc. IEEE Infocom}, 2007.

\bibitem{KM03}
R.~Koetter and M.~M\'{e}dard.
\newblock An algebraic approach to network coding.
\newblock {\em {IEEE/ACM} Trans. Networking}, 11(5):782--795, 2003.

\bibitem{LSB06}
M.~Langberg, A.~Sprintson, and J.~Bruck.
\newblock The encoding complexity of network coding.
\newblock {\em {IEEE} Trans. Inform. Theory}, 52(6):2386--2397, 2006.

\bibitem{LYC03}
S.-Y.~R. Li, R.~W. Yeung, and N.~Cai.
\newblock Linear network coding.
\newblock {\em {IEEE} Trans. Inform. Theory}, 49(2):371--381, 2003.

\bibitem{RP86}
M.~B. Richey and R.~G. Parker.
\newblock On multiple {S}teiner subgraph problems.
\newblock {\em Networks}, 16(4):423--438, 1986.

\bibitem{TM06}
E.-G. Talbi and H.~Meunier.
\newblock Hierarchical parallel approach for {GSM} mobile network design.
\newblock {\em J. Parallel Distrib. Comput.}, 66:274--290, 2006.

\end{thebibliography}

\end{document}